\documentclass[lettersize,journal]{IEEEtran}
\usepackage{amsmath,amsfonts}
\usepackage{algorithmic}
\usepackage{algorithm}
\usepackage{array}
\usepackage[caption=false,font=footnotesize,labelfont=rm,textfont=rm]{subfig}
\usepackage{textcomp}
\usepackage{stfloats}
\usepackage{url}
\usepackage{verbatim}
\usepackage{graphicx}
\usepackage{cite}
\usepackage{multirow}
\begin{document}

\title{Two-Factor Authentication Approach Based on Behavior Patterns for Defeating Puppet Attacks}

        %% <-this % stops a space
\author{Wenhao Wang, 
Guyue Li, \IEEEmembership{Member, IEEE}, 
Zhiming Chu,
Haobo Li, \IEEEmembership{Member, IEEE}, 
Daniele Faccio

% \author{Wenhao Wang, Guyue Li,\thanks{Corresponding author: Guyue Li} \IEEEmembership{Member, IEEE}, Haobo Li, \IEEEmembership{Member, IEEE}, Daniele Faccio \thanks{Haobo Li and Daniele Faccio are with the School of Physics and Astronomy, University of Glasgow. (e-mail: haobo.li@glasgow.ac.uk;  daniele.faccio@glasgow.ac.uk).}}%

\thanks{This work has been submitted to the IEEE for possible publication. Copyright may be transferred without notice, after which this version may no longer be accessible.
}
\thanks{Wenhao Wang, Zhiming Chu and Guyue Li are with the School of Cyber Science and Engineering, Southeast University, Nanjing, 210096, China. (e-mail: guyuelee@seu.edu.cn)}
\thanks{Haobo Li and Daniele Faccio are with the School of Physics and Astronomy, University of Glasgow. (e-mail: haobo.li@glasgow.ac.uk;  daniele.faccio@glasgow.ac.uk).}

}

\iffalse
\IEEEpubid{0000--0000/00\$00.00~\copyright~2021 IEEE}
\fi
\maketitle

\begin{abstract}

Fingerprint traits are widely recognized for their unique qualities and security benefits. Despite their extensive use, fingerprint features can be vulnerable to puppet attacks, where attackers manipulate a reluctant but genuine user into completing the authentication process. Defending against such attacks is challenging due to the coexistence of a legitimate identity and an illegitimate intent. In this paper, we propose PUPGUARD, a solution designed to guard against puppet attacks. This method is based on user behavioral patterns, specifically, the user needs to press the capture device twice successively with different fingers during the authentication process. PUPGUARD leverages both the image features of fingerprints and the timing characteristics of the pressing intervals to establish two-factor authentication. More specifically, after extracting image features and timing characteristics, and performing feature selection on the image features, PUPGUARD fuses these two features into a one-dimensional feature vector, and feeds it into a one-class classifier to obtain the classification result. This two-factor authentication method emphasizes dynamic behavioral patterns during the authentication process, thereby enhancing security against puppet attacks. To assess PUPGUARD's effectiveness, we conducted experiments on datasets collected from 31 subjects, including image features and timing characteristics. Our experimental results demonstrate that PUPGUARD achieves an impressive accuracy rate of 97.87\% and a remarkably low false positive rate (FPR) of 1.89\%. Furthermore, we conducted comparative experiments to validate the superiority of combining image features and timing characteristics within PUPGUARD for enhancing resistance against puppet attacks.

\end{abstract}


\section{Introduction}

\IEEEPARstart{F}{ingerprint} traits have become increasingly popular in recent years due to their distinctiveness, reliability, universality, and security. When compared to alternative biometric authentication methods, fingerprint authentication stands out with remarkably low rates of false rejection (FRR) and false acceptance (FAR), making it a more secure option than traditional password-based authentication, which can be susceptible to theft or forgetfulness. Despite holding a substantial share of the global market and finding use in various scenarios \cite{8962186}, fingerprint authentication is not without its inherent flaws, including susceptibility to presentation attacks.

ISO/IEC 30107 defines presentation attack (PA) as ``presentation to the biometric data capture subsystem with the goal of interfering with the operation of the biometric system"\cite{isointernationalorganizationforstandardization_2016_isoiec}. Since PA was proposed, it has received widespread attention, because the implementation cost of creating artificial fingerprints is very low \cite{marrone2021fingerprint}, and the attacker can use many common materials to complete the imitation of the victim's fingerprint, such as silicone \cite{espinoza2011risk}, plasticine \cite{wiehe2004attacking} and thermoplastic materials \cite{espinoza2011vulnerabilities}. Both hardware-based and software-based methods have been proposed to improve the ability of biometric systems to resist such attacks. Hardware-based solutions rely on other biometric characteristics like odor \cite{baldisserra2005fake,9076651} or pulse oximetry \cite{hengfoss2011dynamic} captured by the biometric system while software-based ones utilize extracted image features \cite{9521212}. 

\begin{figure}[h]
\centering
\includegraphics[width=\columnwidth]{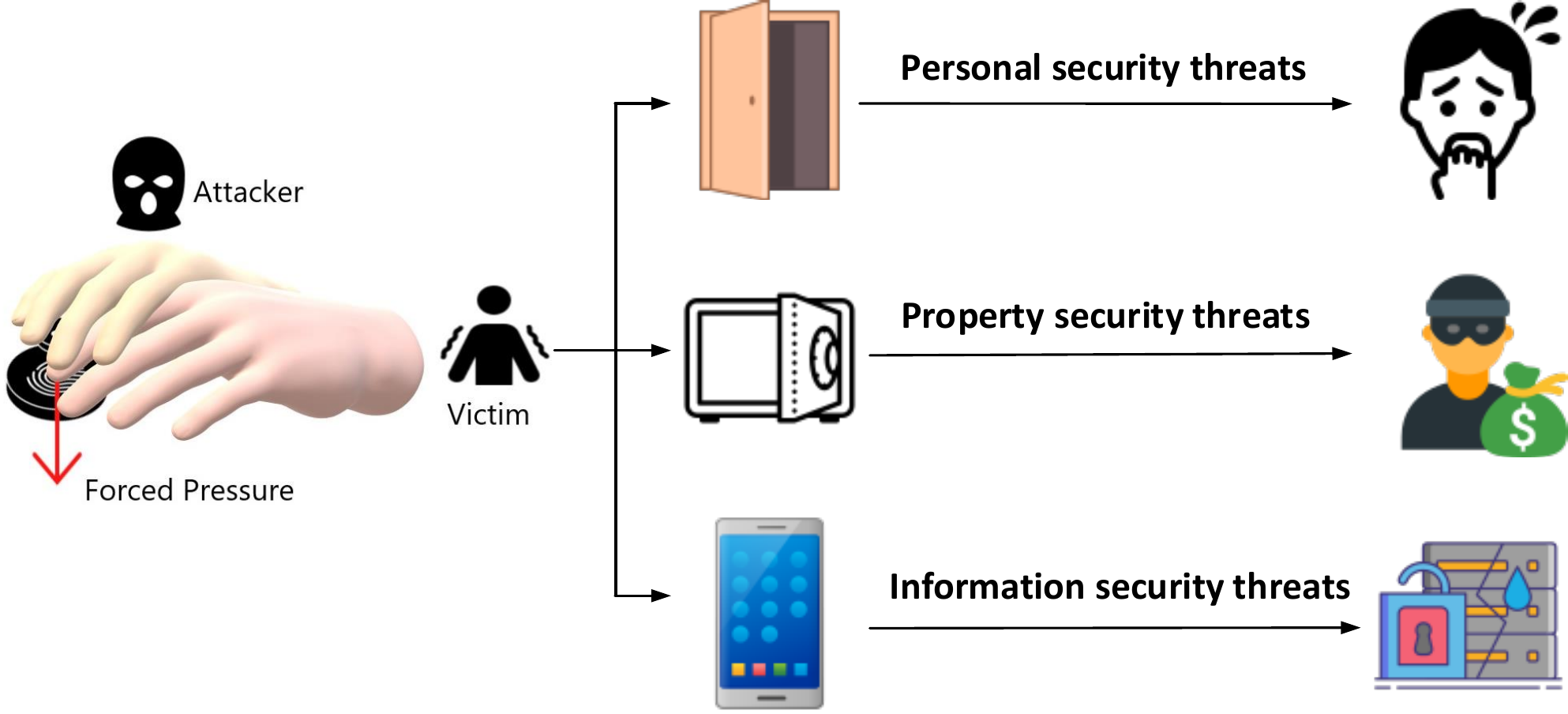}
\caption{Possible security risks caused by puppet attacks.}
\label{risks}
\end{figure}

However, besides detecting fake or altered biometric characteristics, PA also encompasses identifying coercion, non-conformity, and obscuration \cite{sousedik2014presentation}. Puppet attack is an attack in which an attacker forces a legitimate victim to press a finger against a fingerprint reader for intrusion \cite{wu2020liveness}. Puppet attacks often involve violence, threats, or intimidation, such as an attacker wielding a weapon to force a victim to unlock a vault with a fingerprint lock or a child forcibly pressing a parent's finger to unlock a game console. Failing to defend against puppet attacks can result in substantial financial losses and jeopardize personal safety. Hence, it is imperative to research biometric fingerprint authentication methods that can withstand puppet attacks. The schematic diagram of the puppet attack and the security risks it may cause are shown in Fig. \ref{risks}.

Unfortunately, the research on puppet attacks is not as extensive as that on liveness detection. Most of the research on fingerprint presentation attacks focuses on liveness detection, that is, judging whether the input fingerprint comes from a real living person or an imitation. These methods are difficult to defend against puppet attacks, because in puppet attacks, although the victim is coerced, the input fingerprint still belongs to a legitimate user. Wu et al. \cite{wu2020liveness} proposes the concept of puppet attack, in which an attacker places the finger of a legitimate but unwilling victim on the fingerprint acquisition module, %
and designs a detection method based on fingertip touch behavior. However, this method has certain limitations. These include potential false rejection due to behavior variability and different postures, as well as the requirement for the user to hand-hold the device, which can result in failure if the device is placed stationary on a desktop.

In this paper, we introduce PUPGUARD, a solution designed to defend against puppet attacks. 
PUPGUARD leverages user behavior patterns, specifically consecutive finger presses on the fingerprint module using different fingers, to capture intrinsic image features and timing characteristics, and subsequently implements two-factor authentication. This behavior-based approach enhances security by requiring two distinct finger presses and introducing a time gap between them, making it tougher for attackers to mimic the authentication process. Unlike traditional fingerprint authentication, which relies solely on static images, PUPGUARD focuses on dynamic behavior patterns during authentication, strengthening overall security against fingerprint presentation attacks.
We initially conduct separate preprocessing for both fingerprint images and timing characteristics. Subsequently, we employ Local Binary Pattern (LBP), Histogram of Oriented Gradients (HOG) techniques, and Residual Network (ResNet) to extract discriminative features from characterized behavioral patterns. Following this, %we conduct feature fusion of image-based and time-based features to generate a fused feature vector, which is finally input into a one-class classifier to obtain the classification results.
we perform feature selection on image-based features and fuse them with time-based features to create a fused feature vector, which is finally input into a one-class classifier to obtain the classification result.

Based on our investigation, there is currently no publicly available dataset that comprehensively encompasses both image features and timing characteristics required by our PUPGUARD method. 
Specifically, a fingerprint pair is precisely characterized as two distinct fingerprint images acquired through consecutive double presses of the fingerprint module using different fingers during a single authentication process, serving to represent image features.
The corresponding time interval between presses is utilized to represent the timing characteristics.
Existing fingerprint datasets may contain unforced and coerced fingerprint images but do not directly facilitate the formation of fingerprint pairs or the generation of datasets encompassing timing attributes of behavior patterns. This limitation arises from the absence of continuous consecutive presses of the fingerprint module with differing fingers in existing datasets, which fails to reflect the characteristics of continuous pressing in behavior patterns. To address this issue, we established a database comprising 496 fingerprint pairs (992 fingerprints) and corresponding time intervals collected from 31 individuals aged between 20 and 85.

To demonstrate the necessity of our database and the superiority of using PUPGUARD, we conducted a large number of experiments. 
% 实验数据
The results showed that PUPGUARD reaches highest accuracy of 97.87\% and lowest FPR of 1.89\% respectively. The experiment using only image features for detection and the one using only timing characteristics proved the necessity of employing both types of features to represent behavior patterns for detecting puppet attacks. Furthermore, we performed experiments involving behavioral patterns where the same finger was used for two consecutive presses to establish the importance of utilizing two different fingers. Subsequently, we conducted experiments that showed improved performance of PUPGUARD with the expansion of the training set.

The contributions of this paper are summarized as follows:
\begin{itemize}
  \item [1)]
  We propose PUPGUARD, a system that leverages user behavior patterns to capture inherent image features and timing characteristics, thereby implementing a two-factor authentication method. This heightened security approach mandates two separate finger presses with a time gap between them, increasing the difficulty for potential attackers attempting to replicate the authentication process.
  \item [2)]
  To assess the performance of PUPGUARD, we assembled a dataset of 496 fingerprint pairs (comprising 992 individual fingerprints) and their associated time intervals from 31 participants spanning ages 20 to 85. This dataset, obtained with Institutional Review Board (IRB) approval, effectively encapsulates the specified behavioral patterns.
  \item [3)]
  A series of comprehensive experiments were carried out to illustrate both the essentiality and effectiveness of PUPGUARD. These experiments encompassed scenarios using solely image features, exclusively timing characteristics, and employing the same finger for both presses. Our experimental findings conclusively indicate that PUPGUARD attains an outstanding accuracy rate of 97.87\% while simultaneously achieving the lowest false positive rate (FPR) of 1.89\%.
\end{itemize}

The rest of this paper is organized as follows. Section II reviews related work on one-class novelty detection and presentation attack. 
Section III describes the motivation for our work and case studies.
Section IV introduces the data acquisition and preprocessing method to characterize the image features and timing characteristics in PUPGUARD.
Sections V and VI demonstrate feature processing, feature fusion, and classification approaches.
The experimental results and detailed analysis are presented in Section VII. Limitations of PUPGUARD are discussed in Section VIII. Finally, Section VIII provides a summary of this paper.

\section{Related Work}

Fingerprint authentication is susceptible to presentation attacks, as skilled individuals with inexpensive hardware and software can easily generate synthetic fingerprints, thereby increasing their chances of successfully executing such attacks \cite{karampidis2021comprehensive}.

Hardware-based PAD methods necessitate the inclusion of specific sensors within the fingerprint biometric system. These sensors are responsible for verifying the authenticity of signals, such as pulse oximetry \cite{reddy2008new}, blood pressure \cite{drahansky2006liveness}, \cite{lapsley1998anti}, and odor \cite{baldisserra2005fake}. By capturing both the fingerprint and one or more of these signals, the biometric system can authenticate the user. Additionally, some hardware-based techniques involve differentiating between the electrical properties \cite{martinsen2007utilizing,shimamura2011impedance} of living skin and counterfeit materials, as well as utilizing optical coherence tomography (OCT) \cite{cheng2006artificial,bossen2010internal,cheng2007vivo,nasiri2011anti,liu2013capturing}.

Software-based methods use image processing techniques to extract image features from acquired images, combined with machine learning methods to improve defense against fingerprint spoofing attacks \cite{wu2021toward}. Specifically, software-based methods can be divided into dynamic and static methods. Dynamic techniques utilize time-varying features that require a sequence of fingerprint images or videos to extract \cite{karampidis2021comprehensive}. These features identify the authenticity of fingerprints by detecting the physiological characteristics of the human body. Current mainstream methods include skin distortion-based methods \cite{antonelli2006fake,zhang2007fake,jia2007new} and perspiration-based methods \cite{derakhshani2003determination,abhyankar2009integrating,marcialis2010analysis,memon2011active}. Unlike dynamic methods, static methods only need one image of the fingerprint. They extract the required features from the image to complete the detection of PA. Methods based on physiological or anatomical features mainly utilize perspiration \cite{tan2010spoofing,johnson2014fingerprint} and sweat pores on the finger surface \cite{espinoza2011using,marasco2012combining,choi2007aliveness}. Methods based on the surface coarseness \cite{pereira2013spatial} of the fingerprint rely on the premise that the surface of the fake fingerprint is rougher \cite{moon2005wavelet} to judge the authenticity of the fingerprint.  Moreover, texture feature based methods are widely employed. Coli et al. \cite{coli2007power} uses high-frequency energy to tell a finger from a fake, because a fake finger does not retain the high-frequency details of a live one. Ghiani et al. \cite{ghiani2012fingerprint} proposed a method based on rotation-invariant local phase quantization, which exploits the lack of information during the fabrication of fake fingerprints and extracts the texture features of fingerprint images to reject fake fingerprints.

Unfortunately, most of the existing researches on presentation attacks focus on liveness detection, so it is difficult for these methods to detect puppet attacks. Existing methods of defending against puppet attacks have certain flaws. Wu et al. \cite{wu2021toward} introduces the concept of puppet attack and designs a detection method based on fingertip-touch behavior. However, this method requires the user to hand-hold the authentication device and the need for a handheld authentication device makes it difficult to apply the method to scenarios where the fingerprint device is stationary, such as a door lock or safe. Therefore, a method that can authenticate both when the user is holding the authentication device and when the device is stationary is needed to fill the gap of current research in usage scenarios. Our proposed PUPGUARD will be developed towards this goal while guaranteeing high accuracy and low false positive rate.

\section{Principle of PUPGUARD}

We represent a legitimate user experiencing a puppet attack as a combination of two attributes: the user's genuine identity and an illegitimate state. The concurrent presence of these two attributes is what complicates the defense against puppet attacks. To successfully counter such attacks, it becomes essential to identify and discern these two attributes during the user authentication process. If we consider these two attributes as Boolean values and view puppet attack detection as the logical ``and" relationship between them, then the user is deemed legitimate only when both attributes hold true -- meaning the user possesses a legitimate identity and a legitimate state.

Conventional fingerprint authentication methods commonly employ a scheme where the user presses the fingerprint acquisition module once, and the classifier determines the legitimacy of the user's identity based on this static fingerprint image. These approaches pose challenges in identifying the state attribute of a puppet attack because, even during an attack, the fingerprint image captured by the device remains that of the legitimate user.
Therefore, extracting the state attributes of the user authentication process is the key to PUPGUARD's defense against puppet attacks.

We are aware that when an individual's state becomes abnormal, it frequently manifests through specific behavioral patterns, such as trembling, stiffness, weakness, or the use of excessive force. In situations where a user is subjected to a puppet attack and compelled to undergo authentication against their will, the victim's response can vary from resistance due to anger, trembling due to fear, to stiffness and powerlessness due to disorientation. Consequently, in PUPGUARD, our emphasis is on analyzing the user's behavioral patterns to extract the state-related characteristics of the authentication process, facilitating the detection of puppet attacks.

As mentioned earlier, the conventional approach of static fingerprint image detection, based on a single press, poses challenges in extracting user state attributes. Therefore, in the context of PUPGUARD, we focus on the authentication process in which the user presses the fingerprint module twice. In this authentication procedure, the user is required to consecutively press the capture module twice using different fingers, and we classify this sequential behavior as a behavioral pattern within PUPGUARD. The necessity of utilizing distinct fingers for these two presses will be explored further in Section V in correlation with the experiments. We can break down this behavioral pattern into a series of progressively executed actions, which include pressing with the first finger, switching fingers, and then pressing with the second finger. Notably, the presence of a finger-switching action between these two presses indicates the existence of a non-negligible time interval. In perceptual terms, when a user is under attack, resistance or trembling, to some extent, prolongs the time the attacker compels the victim to align their finger with the fingerprint module. This, in turn, extends the duration needed to switch fingers between the two presses.
As demonstrated below, switching fingers while under attack takes much longer compared to the normal state, indicating an abnormal behavioral pattern of the user during the authentication process, which in turn indicates an abnormal state, i.e., under attack.

In the following, we first prove that this time interval for switching fingers is measurable; then we show that the victim's behavioral pattern is quite different from the normal state when under attack; and finally we demonstrate the framework of PUPGUARD.

\begin{figure}[ht]
\centering
\includegraphics[width=86mm]{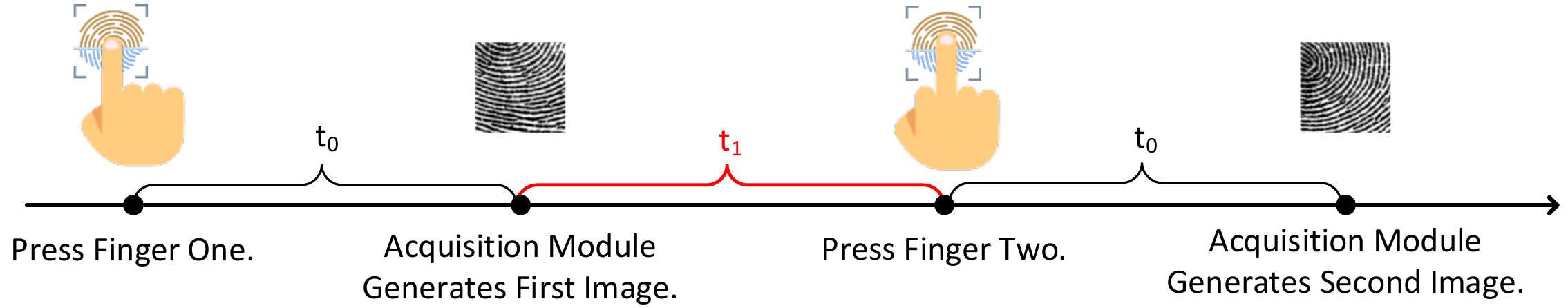}
\caption{Measurable time difference.}
\label{interval}
\end{figure}

\subsection{Why this time interval can be measured accurately}
% As shown in Fig. \ref{interval}, we put the whole process on a time axis, and the four important stages have been marked on it. 
As shown in Fig. \ref{interval}, we put the whole process of the defined behavioral pattern on a timeline and marked four important stages on it.
After the user finishes pressing the finger, the acquisition device will complete image generation after $t_0$. This $t_0$ is completely determined by the hardware performance of the fingerprint collection device and has nothing to do with the user. Therefore, no matter whether Finger 1 or Finger 2 is pressed, the device completes image generation after $t_0$. It can be seen from the figure that it takes $t_0+t_1$ for the user to switch fingers, and the time difference between two image generation by the acquisition device is exactly $t_0+t_1$. In other words, the time it takes for the user to switch fingers is exactly the same as the time it takes for the device to complete the two actions. Therefore, although the time difference between the user switching fingers is difficult to measure, we can easily measure the time difference between two operations completed by the hardware device. Under the same hardware conditions, this time difference is completely driven by the user's behavioral habits and the state during the fingerprint presses.

\begin{figure}[ht]
\centering
\subfloat[Successfully press the acquisition device.]{\includegraphics[width=40mm,height=40mm]{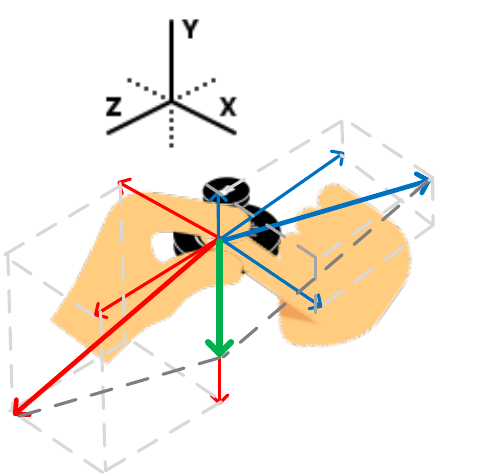}%
\label{force1}}
\quad
\quad
\quad
\quad
\quad
\subfloat[Slip away from the acquisition device.]{\includegraphics[width=41mm,height=41mm]{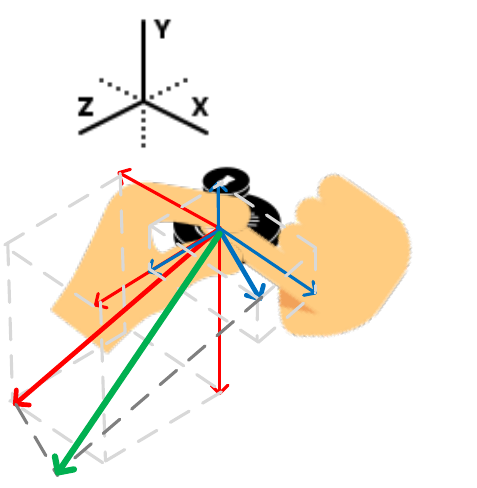}%
\label{force2}}
\caption{Force analysis in two cases.}
\label{force_analysis}
\end{figure}

% 为什么行为模式可以有效反映用户状态
\subsection{Why Behavioral Patterns are Effective in Reflecting User States}
Through the above analysis, we already know that the defined behavioral patterns can be accurately captured by the hardware, and in more detail, the fingerprint images of two presses will be captured by the sensor, and the time interval between two presses can be accurately measured by recording the generation time of two images. In the following, we analyze in detail the differences between the defined behavioral patterns when the user is in a normal state and when subjected to a puppet attack.

When a user completes authentication normally, he or she presses the fingerprint sensor at the rate, direction, and force to which he or she is accustomed, and the switching of fingers between presses is natural and consistent. However, when the victim is forced by the attacker to align the finger with the sensor, the victim's behavioral pattern shows a huge difference compared to the normal state. We explain such a reason by analyzing the forces in two pressing scenarios. As shown in Fig. \ref{force_analysis}, the attacker's force is shown in red arrows, the victim's force is shown in blue arrows, and the resultant force is shown in green arrows. At this moment in Fig. \ref{force_analysis}(a), the magnitude of the forces in the x and z directions are equal but opposite for the attacker and the victim, while in the y direction, the force exerted by the victim is smaller than the force exerted by the attacker pressing down, so the resultant force is downwards and the attacker can force the victim to press the fingerprint acquisition module. However, as shown in Fig. \ref{force_analysis}(b), even when the victim changes the direction of the force applied only in the z-axis, there is a significant change in the direction of the resultant force, which causes the victim's finger controlled by the attacker to deviate from the collection device. 

The above analysis leads us to the following two conclusions, i) no matter how disparate the strength difference between the victim and the attacker is, it is very difficult for the attacker to align the victim's finger to the sensor within the time interval in the normal state, because in the case of the victim struggling and the attacker forcibly controlling it, even a small change of the victim's strength can lead to a significant change of the resulting combined force. ii) resistance movements that may occur in a victim of a puppet attack, such as moving the finger away from the sensor or rotating the finger as far as possible when forced to press, can make the resulting fingerprint image significantly different from that in the normal state, e.g., the center of the press, the angle of rotation, or the force of the press.

Therefore, the above differences in behavioral patterns in the normal state and when under attack is exactly how PUPGUARD can detect puppet attacks.

\begin{figure*}[t]
\centering
\includegraphics[width=\linewidth]{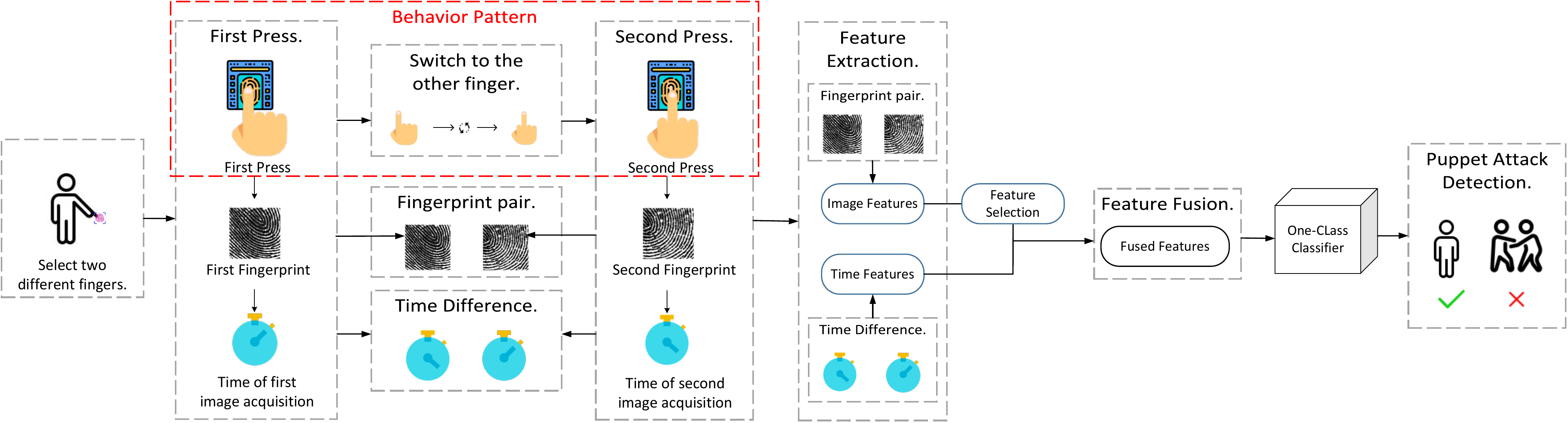}
\caption{Framework of the proposed PUPGUARD.}
\label{framework}
\end{figure*}

\subsection{Framework of PUPGUARD}
The framework of PUPGUARD is shown in Fig.~\ref{framework}. PUPGUARD utilizes user behavior patterns to capture intrinsic image features and timing characteristics, subsequently integrating a two-factor authentication mechanism. This approach bolsters security by necessitating two distinct finger presses and introducing a time gap between them, rendering it more challenging for potential attackers to replicate the authentication procedure.

Our initial process involves the independent preprocessing of both fingerprint images and timing characteristics. Subsequently, we apply Local Binary Pattern (LBP), Histogram of Oriented Gradients (HOG), and Residual Networks (ResNet) to extract distinctive features from characterized behavioral patterns. 
Then we perform feature selection on the image-based features.
Following this, we merge image-based and time-based features through feature fusion, creating a fused feature vector. 
% 值得注意的是，我们也尝试了决策层融合，这会在后续章节中介绍。
This vector is then fed into a one-class classifier to derive the final classification results.
It is worth noting that we also experiment with decision level fusion, which will be presented in subsequent sections.

\section{Proposed Method}
% The framework of PUPGUARD is shown in Fig.~\ref{framework}.
The workflow of PUPGUARD can be divided into the following steps: data acquisition, data preprocessing, feature extraction and selection, feature fusion, and classification. 
% 我们也尝试不使用特征融合，而是对两种特征分别分类，并使用决策融合。
We also try not to use feature fusion but to classify the two features separately and apply decision fusion.
% We also compare the performance of decision level fusion and feature fusion.
%%%%% 在决策时，我们也尝试将两种特征分别使用分类器判别，并对结果进行融合。
Therefore, in this section, we present the implementation details of the above steps one by one.

\subsection{Data Acquisition}
Since the PUPGUARD method requires experimental data derived from a specific behavioral pattern, it is not possible to directly utilize existing databases for experimental data. Here we show the data collection and data acquisition process of PUPGUARD.

\subsubsection{Fingerprint Acquisition Module}
\begin{figure}[t]
\centering
\includegraphics[width=\columnwidth]{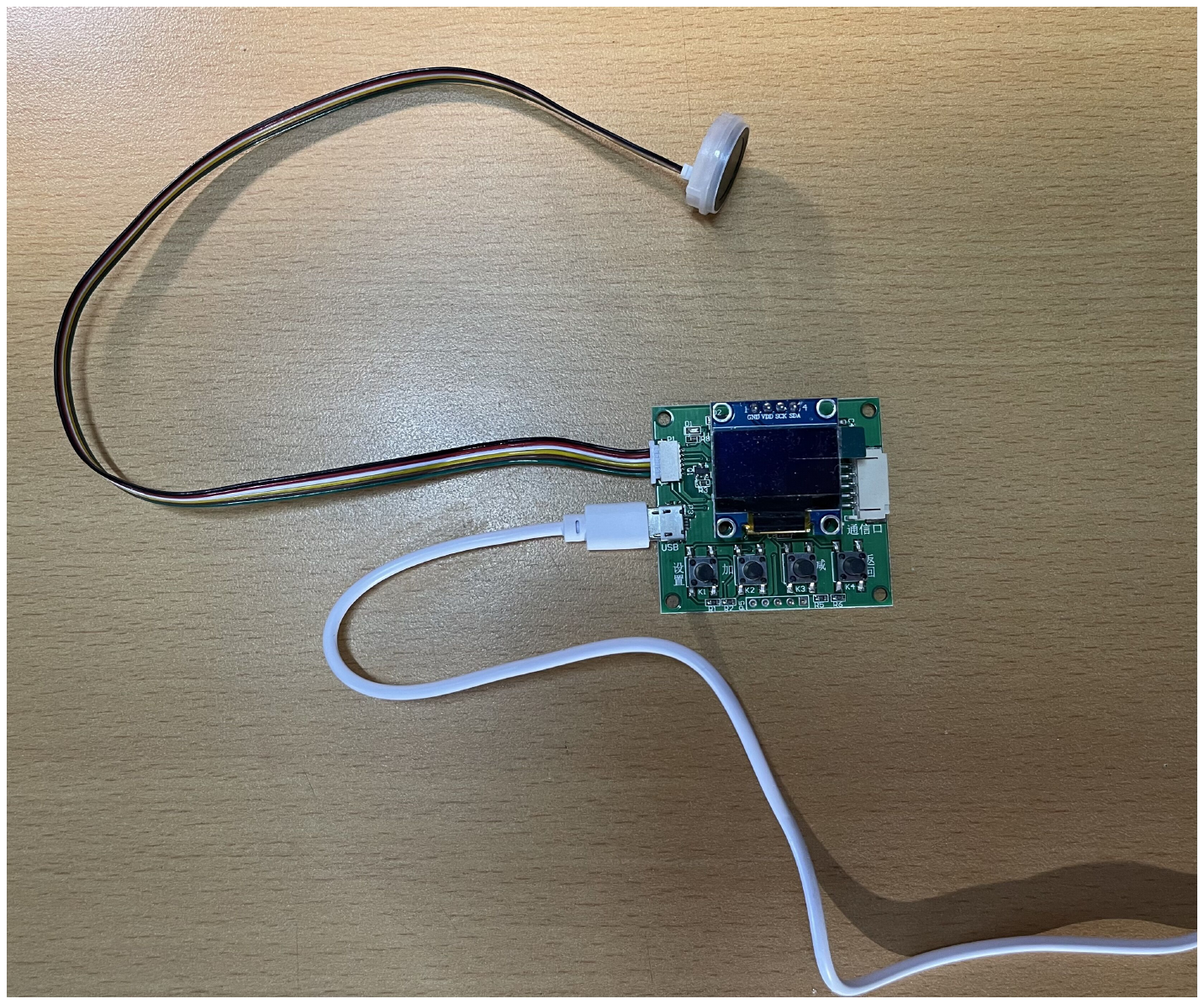}
\caption{Fingerprint acquisition module.}
\label{Sensor_System}
\end{figure}

We compared a variety of fingerprint acquisition modules, and finally chose BM2166 semiconductor fingerprint module, because it integrates semiconductor sensor and fingerprint algorithm chip, and has the advantages of small size, low power consumption, simple interface, high module reliability, and good adaptability to wet and dry fingers. The fingerprint module and STM32 micro-controller together form the fingerprint acquisition system, as shown in Fig. \ref{Sensor_System}. The imaging speed of the system meets our needs for fingerprint acquisition. Specifically, the system can capture fingerprint images in various pressing situations, whether the volunteer is pressing at various angles and centers, or when the volunteer's finger is unintentionally and subtly sliding or rolling during the pressing process. At the same time, the system reads at a satisfactory speed, not too slow to cause a long dataset creation process, nor too fast to cause loss of fingerprint details.

Our research utilized the system for fingerprint extraction. The collected fingerprint image size is 8mm $\times$ 8mm with an image pixel size of 160 $\times$ 160, and a resolution of 508 DPI. The working temperature ranges from $-20^\circ$C to $+40^\circ$C, while the storage temperature ranges from $-40^\circ$C to $+70^\circ$C. Additionally, the working relative humidity is from 40\% to 85\%. At the same time, the system records the current time in standard format \texttt{yyyymmddHHMMSS.xxxxxx} each time it successfully captures a fingerprint image. In this format, \texttt{yyyy} represents the four-digit year, \texttt{mm} represents the two-digit month, \texttt{dd} represents the two-digit day of the month, \texttt{HH} represents the two-digit hour of the day in 24-hour format, \texttt{MM} represents the two-digit minute in the hour, \texttt{SS} represents the two-digit number of seconds in the minute, and \texttt{xxxxxx} represents the six-digit number of microseconds in the second.

\subsubsection{Acquisition Details}
Successful data entry is defined as follows in accordance with the behavioral pattern: volunteers, in a relaxed and natural state, selecting two different fingers and pressing the fingerprint collection module twice in a row, with each finger pressing the module once, in a continuous and natural manner without deliberate pauses or accelerations. 
% 特殊情况
We ensure that all volunteers' pressing actions are considered normal, accommodating various legitimate scenarios that may occur. For instance, if after pressing the first finger, the volunteer notices dust on the second finger, they can simply wipe it off and proceed with the second pressing action. Similarly, if the volunteer encounters any other minor interruptions or adjustments during the process, they can be accommodated as long as they align with the overall requirements of the behavioral pattern.

The pressing gestures of the volunteers on the fingerprint module include pressing with the fingertips, the side of the finger, the middle of the finger, and the bottom of the finger. Since almost all volunteers are not accustomed to using their ring fingers for fingerprint pressing, only 7 volunteers participated in data entry with their ring fingers, completing a total of 31 pairs of fingerprints with the ring finger. Other volunteers were asked to use their thumbs, index fingers, middle fingers, and little fingers to complete the data entry. Each volunteer needed to complete two successful data entries in the following ways: (1) press the thumb first and then the index finger; (2) press the index finger first and then the thumb; (3) press the thumb first and then the middle finger; (4) press the middle finger first and then the thumb; (5) press the index finger first and then the middle finger; (6) press the middle finger first and then the index finger. Each volunteer needed to complete one successful data entry in the following ways: (7) select the thumb and ring finger to complete the data entry; (8) select the thumb and little finger; (9) select the middle finger and ring finger; (10) select the middle finger and little finger. Therefore, each volunteer needed to complete 16 ($4\times2 + 4\times1 = 16$.) successful data entries, or 32 fingerprint images per person. Specifically, the order of (7), (8), (9), and (10) was specified by us. For example, in (7), the volunteer would choose whether to press the index finger first or the middle finger first, and we ensured that the difference in the number of times the two fingers were pressed first would not exceed 1. 

A complete data acquisition process of the acquisition system can be summarized in the following steps: i) the volunteer selects two different fingers, ii) the two fingers are pressed consecutively according to the requirements of a specific behavioral pattern, iii) the system sets up the two captured fingerprint images as a fingerprint pair, iv) the system records the moments of the two fingerprint acquisitions and makes the difference, and v) the system adds the fingerprint pair and the time difference to the dataset as a set of data.

During the data entry process, the collection device was fixed on a table at a height of 1.2 meters. Half of the volunteers needed to stand in front of the collection device to complete the data collection, while the other half needed to sit in front of the collection device. Between each successful data entry behavior, volunteers were required to completely remove their fingers from the fingerprint collection device to ensure a significant difference between each data entry behavior.

During a successful data entry process, when each fingerprint image is successfully entered, the acquisition system will record the current time. The system captures the time difference between the second fingerprint entry and the first fingerprint entry. This time difference serves as the timing characteristics, enabling the detection of puppet attacks.

\subsubsection{Data Constitution}
\begin{figure}[t]
\centering
\subfloat[Different pressing gestures]{
    \begin{minipage}{0.32\linewidth}
		\vspace{2pt}		\centerline{\includegraphics[width=24mm,height=24mm]{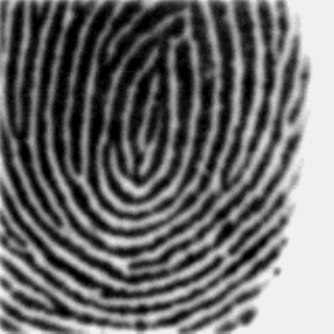}}
	\end{minipage}
	\begin{minipage}{0.32\linewidth}
		\vspace{2pt}		\centerline{\includegraphics[width=24mm,height=24mm]{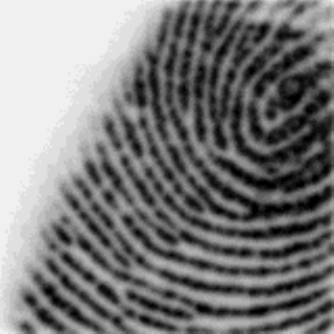}}
	\end{minipage}
	\begin{minipage}{0.32\linewidth}
		\vspace{2pt}		\centerline{\includegraphics[width=24mm,height=24mm]{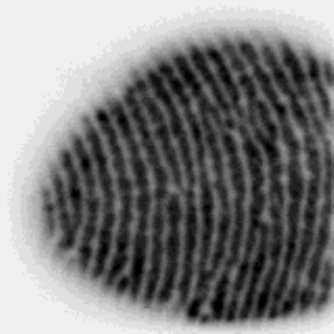}}
	\end{minipage}
 }
 
\subfloat[Different degrees of fingerprint wear.]{
    \begin{minipage}{0.32\linewidth}
		\vspace{2pt}		\centerline{\includegraphics[width=24mm,height=24mm]{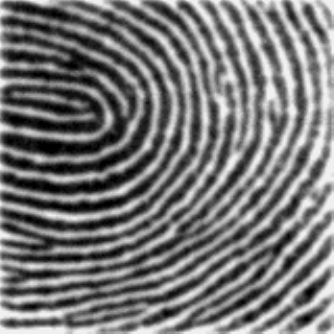}}
	\end{minipage}
	\begin{minipage}{0.32\linewidth}
		\vspace{2pt}		\centerline{\includegraphics[width=24mm,height=24mm]{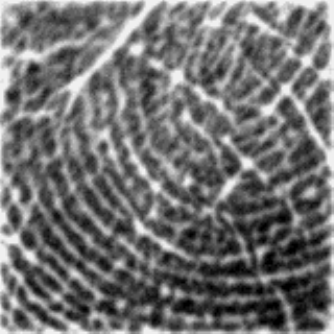}}
	\end{minipage}
	\begin{minipage}{0.32\linewidth}
		\vspace{2pt}		\centerline{\includegraphics[width=24mm,height=24mm]{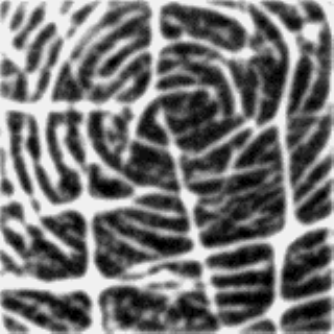}}
	\end{minipage}
 }
\caption{Sample fingerprints in the dataset.}
\label{fingerprint}
\end{figure}

The dataset contains only data collected from volunteers in their normal state, which means that it does not include any anomalous data collected from volunteers who are under puppet attacks. The dataset encompasses various pressing postures that users would naturally adopt, including different pressing angles and centers, as shown in Fig. \ref{fingerprint}(a). Specifically, the press poses completed by volunteers during data entry include pressing with fingertips, pressing with the middle of fingers, pressing with the side of fingers, and pressing with the lower part of fingers. At the same time, the dataset includes various combinations of two presses with different fingers, such as using the thumb first and then the index finger, or using the middle finger first and then the ring finger. The dataset contains 124 fingerprint pairs, which are combinations of thumb and index finger (248 fingerprint images). Combinations refer to pressing two different fingers in two different orders, such as pressing the thumb first and then the index finger, or vice versa. 

The dataset also includes 124 fingerprint pairs of the combination of thumb and middle finger, 124 fingerprint pairs of the combination of index finger and middle finger, 31 fingerprint pairs of the combination of thumb and ring finger, 31 fingerprint pairs of the combination of thumb and little finger, 31 fingerprint pairs of the combination of middle finger and ring finger, and 31 fingerprint pairs of the combination of middle finger and little finger.
% Altogether 31 subjects were involved in the data collection, who were aged from 20 to 85. 12 subjects were female, and 19 were male.
% The dataset we created includes fingerprint images of 31 users aged between 20 and 85 years, including 19 males and 12 females. The gender distribution and age distribution are shown in Fig. \ref{distribution}. 
A total of 31 participants were involved in the data collection process, comprising 12 females and 19 males. Their ages ranged from 20 to 85 years, with 9 participants falling within the 20-30 age range, 6 participants within the 30-45 age range, 6 participants within the 45-50 age range, 7 participants within the 50-56 age range, and 3 participants within the 56-85 age range.
The larger age range ensures that the dataset encompasses the condition of fingerprint wear in all age groups, as shown in Fig. \ref{fingerprint}(b).

\subsection{Data Preprocessing}
The preprocessing of experimental data is divided into two parts: preprocessing of fingerprint images and preprocessing of timing characteristics. For timing characteristics, we standardize them. For fingerprint images, we ultize two different preprocessing methods, one using the classical image segmentation algorithm and the other based on resizing, cropping and normalization.

\subsubsection{Image Preprocessing Based on Otsu}

For fingerprint image segmentation, we employ the Otsu method. Otsu's thresholding algorithm finds a threshold value to separate image foreground and background based on grayscale variance \cite{otsu1979threshold}. This robust technique handles varying lighting, contrast, and noise levels in image processing tasks.

Given an image with L gray levels and pixel count $n_i$ for gray value $i$, the total pixel count $N$ is:

\begin{equation}
    N = \sum_{i=0}^{L-1} n_i
\end{equation}

The pixel probability $p_i$ for gray value $i$ is:

\begin{equation}
    p_i = \frac{n_i}{N}
\end{equation}
where $p_i\geq0$, $\sum_{i=0}^{L-1} p_i=1$.

The mean gray value of the whole image is:
\begin{equation}
    m=\sum_{i=0}^{L-1} i p_i
\end{equation}

Defining threshold $k$ to divide pixels into classes $C_1$ and $C_2$ with probabilities $P_{C_1}(k)$ and $P_{C_2}(k)$, the mean gray values of these classes are:

\begin{equation}
    m_{C_1} = \frac{1}{P_{C_1}(k)} \sum_{i=0}^{k} i p_i
\end{equation}

\begin{equation}
    m_{C_2} = \frac{1}{P_{C_2}(k)} \sum_{i=k+1}^{L-1} i p_i
\end{equation}

The between-class variance is:

\begin{equation}
    \begin{aligned}
        \sigma_B^2 (k)&=P_{C_1}  (m_{C_1} - m)^2 + P_{C_2}  (m_{C_2} - m)^2\\
        &=P_{C_1}  P_{C_2}  (m_{C_1} - m_{C_2})^2\\
        &=\frac{(m P_{C_1}-\sum_{i=0}^{k} i  p_i)^2}{P_{C_1}(1-P_{C_1})}
    \end{aligned}
\end{equation}

The optimal threshold $k^{*}$ maximizes $\sigma_B^2 (k)$:

\begin{equation}
    \sigma_{B}^{2}\left(k^{*}\right) = \max_{0 \leq k \leq L-1} \sigma_{B}^{2}(k)
\end{equation}

\begin{figure}[htbp]
\centering
\subfloat[Original fingerprint image]{\includegraphics[width=24mm,height=24mm]{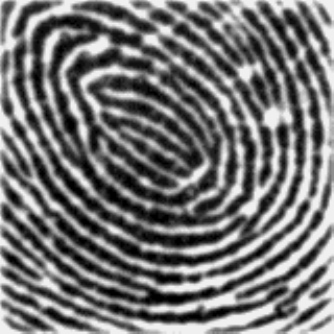}%
\label{ori}}
\quad
\quad
\quad
\quad
\quad
\subfloat[Fingerprint image using Otsu]{\includegraphics[width=24mm,height=24mm]{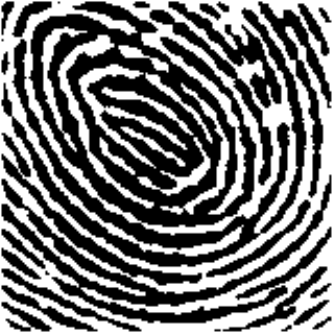}%
\label{otsu}}
\caption{Comparison of fingerprint images before and after using Otsu.}
\label{ori and otsu}
\end{figure}

Utilizing this optimal threshold $k^{*}$ achieves image segmentation. To visualize, Fig. \ref{ori and otsu} contrasts the original and Otsu processed images. This preprocessing approach is labeled \textit{Prepro1}.

\subsubsection{Image Preprocessing Based on Resizing, Cropping, and Normalization}
The images in our training dataset undergo a series of preprocessing steps to prepare them for analysis. Initially, these images are subjected to resizing and center cropping to achieve uniformity in size, ensuring that they can be effectively processed by our model. Subsequently, we convert the images into PyTorch tensors, as this format is compatible with our chosen model architecture.

Once the images are transformed into tensors, we take an essential step in the preprocessing pipeline, which involves normalizing the pixel values. This normalization process is crucial for achieving standardized data representation throughout the subsequent processing stages. By scaling the pixel values appropriately, we bring the images to a common scale and remove any potential biases in the data.

The combination of resizing, center cropping, converting to tensors, and pixel value normalization forms a critical foundation for the success of our model during training. These preprocessing steps allow the model to effectively learn and extract meaningful features from the images, leading to better performance and generalization on unseen data. This preprocessing approach is labeled \textit{Prepro2}.

\subsubsection{Timing Characteristics Standardization}

For timing characteristics standardization, we utilize the formula:

\begin{equation}
    t^{*} = \frac{t - \mu}{\sigma}
\end{equation}

where $\mu$ is the mean and $\sigma$ is the standard deviation of the sample data.

\subsection{Feature Extraction and Feature Selection}
In this subsection, feature extraction and feature selection is discussed. Since timing characteristics is one-dimensional data, normalized timing data is directly used as timing Characteristics. For the preprocessed fingerprint images, we use and compare two different features, i.e., LBP and HOG based features, and residual network (ResNet) based features. 
% 为了选择最佳的图像特征组合，并减少特征维度，我们也对图像特征进行了特征选择。
To select the best feature combinations as well as reduce the feature dimensions, we also perform feature selection on the image features.

\subsubsection{LBP- and HOG-Based Features}
The Local Binary Pattern (LBP) algorithm, which was first proposed by Ojala et al. in 1994 for texture classification \cite{ojala1994performance}, is a widely used texture descriptor in computer vision applications. The LBP operator works by comparing the intensity values of each pixel with its neighboring pixels within a local region, typically a $3 \times 3$ or $5 \times 5$ window. For each pixel, a binary code is assigned based on whether the neighbor's intensity is greater or less than the central pixel's intensity. This binary code is then used to generate a histogram of the texture pattern within the region. Let $p$ be the central pixel of a local region, and $q$ be a neighboring pixel. Then, the binary code for $q$ is defined as:
\begin{equation}
    B(q)=
    \begin{cases}
        1   & \text { if } q\geq p \\
        0   & \text { if } q < p
    \end{cases}
\end{equation}

The LBP code for $p$ is then calculated by concatenating the binary codes for all neighboring pixels in a clockwise order. For example, a $3 \times 3$ window would have 8 neighboring pixels, and the LBP code would be a concatenation of their binary codes, starting from the pixel to the right of $p$ and moving clockwise around the window. Finally, a histogram is constructed by counting the occurrences of each unique LBP code within the local region. This histogram can then be used as a texture descriptor for further analysis.

The Histogram of Oriented Gradients (HOG) algorithm works by analyzing the gradient orientations of small image patches and constructing histograms of these orientations. These histograms are then normalized and concatenated to form a feature vector that represents the image. More specifically, first of all, compute gradient images in $x$ and $y$ directions using a filter such as Sobel. Then, compute gradient magnitudes and orientations for each pixel in the image as follows:
\begin{equation}
    G(x, y) = \sqrt{G_x(x,y)^2 + G_y(x,y)^2}
\end{equation}
% \begin{equation}
%     G_x(x,y)=I(x+1, y) - I(x-1, y)
% \end{equation}
% \begin{equation}
%     G_y(x,y)=I(x, y+1) - I(x, y-1)
% \end{equation}
\begin{equation}
    \theta (x, y) = \arctan{\frac{G_y(x,y)}{G_x(x,y)}}
\end{equation}
where %$I(x, y)$ is the intensity of the image at pixel $(x, y)$, 
$G_x(x,y)$ is the gradient along the $x$ direction, $G_y(x,y)$ is the gradient along the $y$ direction. After that, divide the image into cells of a fixed size (e.g. $8 \times 8$ pixels). For each cell, create a histogram of gradient orientations weighted by gradient magnitudes. Then combine adjacent cells into larger blocks (e.g. $2 \times 2$ cells). Normalize the histograms in each block to account for variations in lighting and contrast. Finally, concatenate the histograms from all blocks into a single feature vector.

% \subsection{Neural Network-Based Features}
\subsubsection{Residual Network-Based Features}
ResNet, short for Residual Network, is a deep convolutional neural network architecture proposed by Kaiming He et al. in 2015 \cite{he2016deep}. It utilizes residual blocks, employing ``skip connections" to pass residual information, effectively tackling the vanishing gradient problem in deep networks. ResNet allows the construction of very deep networks and achieves outstanding performance in computer vision tasks.

\begin{figure}[]
\centering
\includegraphics[width=\columnwidth]{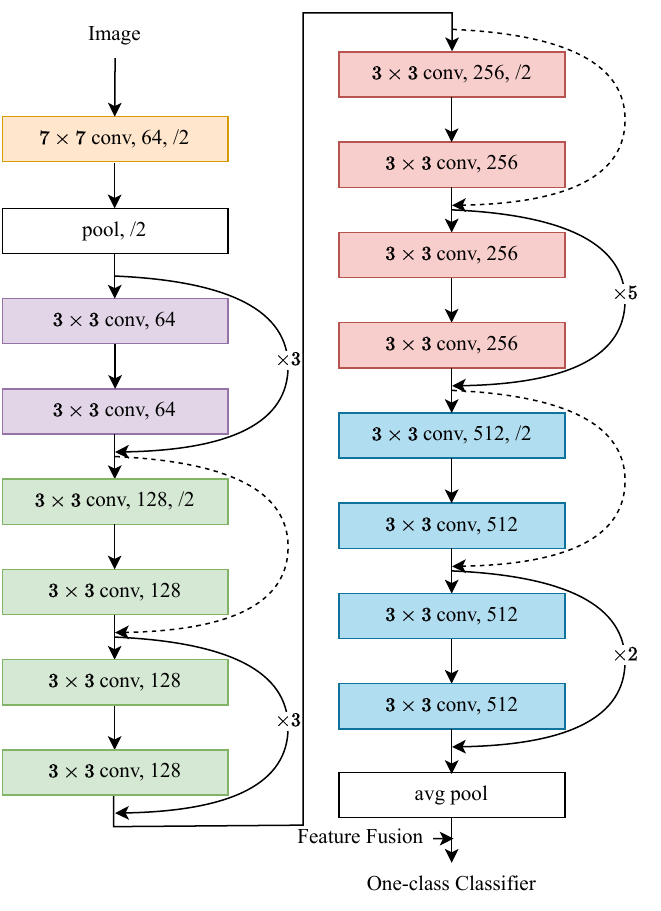}
\caption{Framework of ResNet34-based Feature Extractor.}
\label{resnet34_occlassifier}
\end{figure}

To leverage ResNet for image feature extraction, we perform a modification on the original architecture by discarding the fully connected layer. By doing so, we retain the convolutional and pooling layers, which are responsible for learning hierarchical spatial features, while discarding the classification-specific component. This alteration facilitates the extraction of higher-level, semantically rich feature representations from the input images, which can be utilized for puppet attack detection. For instance, the framework of using the ResNet34 extract features for subsequent classification is shown in Fig. \ref{resnet34_occlassifier}.

\subsubsection{Feature Selection on Image Features}
% PCA
% 使用上面介绍的方法提取图像特征后，相比一维的时间特征，图像特征依然是高维特征。
After extracting the image features using the method described above, the image features are still high dimensional compared to the one-dimensional timing characteristics.
To select the best feature combinations as well as reduce the feature dimensions, we perform feature selection on the image features, and in our experiments we employ Principal Component Analysis (PCA).

PCA is a popular data analysis technique for handling high-dimensional datasets. It achieves dimensionality reduction by linearly transforming data into a new coordinate system while retaining as much information as possible in lower dimensions, thereby enhancing data interpretability. %PCA finds widespread applications across various domains, including genetics, microbiology, and atmospheric science, for dimensionality reduction and exploratory data analysis. 
It identifies principal components, with the first principal component being the direction that maximizes the variance of the projected data, and subsequent principal components being orthogonal to the previous ones while also maximizing the variance of the projected data.

After feature selection and feature dimensionality reduction, the image features will complete feature fusion with one-dimensional timing characteristics, as will be described below.

\subsection{Feature Fusion and Decision Fusion}
% \subsection{Feature Fusion}
In our defined behavior pattern, timing characteristics are represented as one-dimensional, while image features belong to high-dimensional space. 
Therefore, we try two fusion methods to deal with these two features. The first method is feature fusion, where we fuse the two features to form a one-dimensional feature vector, and this fused feature vector can characterize the behavioral patterns more effectively. The second method is decision level fusion, where we use two classifiers, as will be described in the next subsection, to process image features and timing characteristics separately, and then the outputs of the two classifiers are fused to obtain the final classification results.

\subsubsection{Feature Concatenation} 
We concatenate image features and timing characteristics into a single larger feature vector, then use this merged vector for prediction.

\subsubsection{Feature Cross} 
We intersect image features with timing characteristics to generate new combined features. In particular, we multiply each element of the image features with the timing characteristics to create a new feature vector. This method is suitable when there is some correlation between image and time features.

\subsubsection{Decision Level Fusion}
% 除了使用特征融合，我们也尝试使用两个分类器分别处理PUPGUARD中的两类特征，再使用决策融合处理两个分类器的分类结果，得到最终检测结果。
In addition to using feature fusion, we also try to use two classifiers to process the two types of features in PUPGUARD separately, and then use decision fusion to process the classification results of the two classifiers to get the final detection results.

Decision level fusion involves merging decisions or classifications from various sensors or modalities to create a single, robust decision, with the aim of boosting system performance. Its primary objectives are to reduce uncertainty, enhance decision accuracy, and improve reliability by amalgamating information from diverse sources. This typically involves assessing the contribution of each source and applying appropriate weighting for well-balanced information integration. 

% 在我们的实验中,我们尝试使用两个单类分类器分别判别PUPGUARD中的图像特征和时间特征是否属于正常值,之后,我们对两个分类器返回的分类结果进行逻辑与,即当且仅当两个分类器的判断结果都为合法用户时,最终的决策是用户合法.
In our experiments, we try to use two one-class classifiers to discriminate whether the image features and timing characteristics in PUPGUARD are normal values or not, respectively.
% the logical ``and" relationship
After that, we use the logical ``and" relationship to process the classification results returned by the two classifiers, and the final decision is that the user is legitimate if and only if the results of both classifiers are normal.

\subsection{Detection Based on One-class Classifiers}
Since our dataset contains only legitimate user data and no outlier data, this is a one-class classification problem. Therefore, we use the following three models to detect puppet attacks:
i) one-class support vector machine (OC-SVM)
ii) isolation forest (IF)
and iii) local outlier factor (LOF)

\subsubsection{OC-SVM} 
OC-SVM is a type of support vector machine algorithm that is used for novelty detection. The goal of One-class SVM is to learn a decision boundary that separates the normal data points from the outliers. The algorithm takes a single class of input data, typically representing the normal class, and learns a decision boundary that maximizes the margin around the normal data points \cite{scholkopf1999support}. This margin is defined as the distance between the decision boundary and the closest data point from the normal class.

\subsubsection{LOF} 
LOF is based on the concept of local density, determined by considering k nearest neighbors and their distances \cite{breunig2000lof}. By comparing the local density of an object with that of its neighbors, regions with similar density can be identified, along with points that have significantly lower density than their neighbors, classifying them as outliers.
The local density is estimated by the typical distance at which a point can be ``reached" from its neighbors. The definition of ``reachability distance" used in LOF is an additional measure to produce more stable clustering results.

\subsubsection{IF} 
IF is a popular anomaly detection algorithm introduced by Liu et al\cite{liu2008isolation}. It efficiently identifies outliers in large-scale datasets by creating random binary trees and measuring the isolation of anomalies based on their shorter path lengths from the root. Its non-parametric nature, computational efficiency, and effectiveness in high-dimensional data have made it widely utilized in various domains, including cybersecurity, fraud detection, and fault diagnosis.

\section{Experiments and Analyses}
\subsection{Experimental Preparation and Evaluation Indexes}
% Since our dataset contains only positive samples and no negative ones, 
% 为了评估PIPGUARD的性能
To evaluate the performance of PIPGUARD,
we create a testset that contains 94 fingerprint pairs (188 fingerprint images) and corresponding time difference data, including 41 positive samples and 53 negative samples. 
Abnormal behavior is defined as any instance or combination of the following behaviors during the data collection process: (1) forcefully pressing the fingerprint module with a single finger, (2) forcefully pressing the fingerprint module with both fingers simultaneously, and (3) exhibiting an unusually prolonged or shortened time difference between the two finger presses.

We collected the testset by involving different combinations of male victims and male attackers, female victims and male attackers, male victims and female attackers, and female victims and female attackers. During these experiments, attackers employed various methods to coerce victims into completing the fingerprint pressing, resulting in victims exhibiting abnormal behavior.

We measure the performance of our proposed method with accuracy, FPR, recall, precision and F1-score. Accuracy is the proportion of correct predictions, recall is the probability of correctly predicting positive samples, precision refers to the proportion of correct predictions among all predicted positive samples, FPR is the probability of predicting an abnormal data as normal, and F1-score is the harmonic mean of precision and recall. The mathematical expression of these indicators is as follows:
\begin{equation}
    Accuracy=\frac{TP+TN}{TP+TN+FP+FN}
\end{equation}
\begin{equation}
    FPR=\frac{FP}{FP+TN}
\end{equation}
\begin{equation}
    Recall=\frac{TP}{TP+FN}
\end{equation}
\begin{equation}
    Precision=\frac{TP}{TP+FP}
\end{equation}
\begin{equation}
    F1\text{-}score=2 \times \frac{Precision \times Recall}{Precision+Recall}
\end{equation}

where TP, FP, TN and FN represent the number of true positives, false positives, true negatives and false negatives, respectively.

% In real life, we would rather be shut out of the fingerprint authentication system ourselves than have the system identify an attacker as a legitimate user. In other words, FPR is a very important indicator to measure our defense against puppet attacks, so we need to increase the accuracy as much as possible while reducing the FPR.

% 在实际应用中，被拒之门外相比遭受非法入侵更加令人接收。因此在考察PUPGUARD的表现时，我们需要着重考虑准确率和假阳性率。
In practical applications, being rejected is more acceptable than suffering from illegal intrusion. Therefore, when examining the performance of PUPGUARD, we need to focus on the accuracy and false positive rates.

\begin{table*}[]
\centering
\renewcommand\arraystretch{1.2}
% \setlength{\abovecaptionskip}{0pt}%    
% \setlength{\belowcaptionskip}{10pt}%
% \caption{Experimental Results of PUPGUARD under Different Conditions}
\caption{Experimental Results of PUPGUARD}
\begin{tabular}{|c|c|l|c|c|c|c|c|}
\hline
\textbf{Features}                        & \textbf{One-class Classifier}    & \textbf{Feature Fusion} & \textbf{Accuracy} & \textbf{FPR}    & \textbf{Recall}  & \textbf{Precision} & \textbf{F1-score} \\ \hline
\multirow{6}{*}{ResNet34-based}          & \multirow{2}{*}{OC-SVM}          & Feature Cross           & 93.62\%           & 3.77\%          & 90.24\%          & 94.87\%            & 0.92              \\ \cline{3-8} 
                                         &                                  & Feature Concatenation   & 65.96\%           & 52.83\%         & 90.24\%          & 56.92\%            & 0.70              \\ \cline{2-8} 
                                         & \multirow{2}{*}{IF}              & Feature Cross           & 93.62\%           & 3.77\%          & 90.24\%          & 94.87\%            & 0.92              \\ \cline{3-8} 
                                         &                                  & Feature Concatenation   & 79.79\%           & 32.08\%         & 95.12\%         & 69.64\%            & 0.80              \\ \cline{2-8} 
                                         & \multirow{2}{*}{LOF}             & Feature Cross           & 93.62\%           & 9.43\%          & 97.56\%         & 88.89\%            & 0.93              \\ \cline{3-8} 
                                         &                                  & Feature Concatenation   & 52.13\%           & 84.91\%         & 100.00\%         & 47.67\%            & 0.65              \\ \hline
\multirow{6}{*}{\textbf{ResNet50-based}} & \multirow{2}{*}{\textbf{OC-SVM}} & \textbf{Feature Cross}  & \textbf{97.87\%}  & \textbf{1.89\%} & \textbf{97.56\%} & \textbf{97.56\%}   & \textbf{0.98}     \\ \cline{3-8} 
                                         &                                  & Feature Concatenation   & 68.09\%           & 56.60\%         & 100.00\%          & 57.75\%            & 0.73              \\ \cline{2-8} 
                                         & \multirow{2}{*}{IF}              & Feature Cross           & 93.62\%           & 7.55\%          & 95.12\%         & 90.70\%            & 0.93              \\ \cline{3-8} 
                                         &                                  & Feature Concatenation   & 63.83\%           & 54.72\%         & 87.80\%          & 55.38\%            & 0.68              \\ \cline{2-8} 
                                         & \multirow{2}{*}{LOF}             & Feature Cross           & 88.30\%           & 18.87\%         & 97.56\%          & 80.00\%            & 0.88              \\ \cline{3-8} 
                                         &                                  & Feature Concatenation   & 48.94\%           & 88.68\%         & 97.56\%          & 45.98\%            & 0.63              \\ \hline
\multirow{6}{*}{ResNet101-based}         & \multirow{2}{*}{OC-SVM}          & Feature Cross           & 89.36\%           & 5.66\%          & 82.93\%          & 91.89\%            & 0.87              \\ \cline{3-8} 
                                         &                                  & Feature Concatenation   & 65.96\%           & 49.06\%         & 85.37\%          & 57.38\%            & 0.69              \\ \cline{2-8} 
                                         & \multirow{2}{*}{IF}              & Feature Cross           & 94.68\%           & 3.77\%          & 92.68\%          & 95.00\%            & 0.94              \\ \cline{3-8} 
                                         &                                  & Feature Concatenation   & 76.60\%           & 37.74\%         & 95.12\%          & 66.10\%            & 0.78              \\ \cline{2-8} 
                                         & \multirow{2}{*}{LOF}             & Feature Cross           & 96.81\%           & 5.66\%          & 100.00\%         & 93.18\%            & 0.96              \\ \cline{3-8} 
                                         &                                  & Feature Concatenation   & 53.19\%           & 83.02\%         & 100.00\%         & 48.24\%            & 0.65              \\ \hline
\multirow{6}{*}{ResNet152-based}         & \multirow{2}{*}{OC-SVM}          & Feature Cross           & 90.43\%           & 3.77\%          & 82.93\%          & 94.44\%            & 0.88              \\ \cline{3-8} 
                                         &                                  & Feature Concatenation   & 61.70\%           & 54.72\%         & 82.93\%          & 53.97\%            & 0.65              \\ \cline{2-8} 
                                         & \multirow{2}{*}{IF}              & Feature Cross           & 93.62\%           & 7.55\%          & 95.12\%          & 90.70\%            & 0.93              \\ \cline{3-8} 
                                         &                                  & Feature Concatenation   & 73.40\%           & 43.40\%         & 95.12\%          & 62.90\%            & 0.76              \\ \cline{2-8} 
                                         & \multirow{2}{*}{LOF}             & Feature Cross           & 94.68\%           & 7.55\%          & 97.56\%         & 90.91\%            & 0.94              \\ \cline{3-8} 
                                         &                                  & Feature Concatenation   & 48.94\%           & 90.57\%         & 100.00\%         & 46.07\%            & 0.63              \\ \hline
\multirow{6}{*}{LBP}                     & \multirow{2}{*}{OC-SVM}          & Feature Cross           & 88.29\%           & 15.09\%         & 92.68\%          & 82.61\%            & 0.87              \\ \cline{3-8} 
                                         &                                  & Feature Concatenation   & 43.62\%            & 96.23\%         & 95.12\%          & 43.33\%            & 0.60              \\ \cline{2-8} 
                                         & \multirow{2}{*}{IF}              & Feature Cross           & 45.74\%           & 94.34\%         & 97.56\%          & 44.44\%            & 0.61              \\ \cline{3-8} 
                                         &                                  & Feature Concatenation   & 44.68\%           & 96.23\%         & 97.56\%          & 43.96\%            & 0.61              \\ \cline{2-8} 
                                         & \multirow{2}{*}{LOF}             & Feature Cross           & 79.79\%           & 33.96\%         & 97.56\%          & 68.97\%            & 0.81              \\ \cline{3-8} 
                                         &                                  & Feature Concatenation   & 44.68\%           & 96.23\%         & 97.56\%          & 43.96\%            & 0.61              \\ \hline
\multirow{6}{*}{HOG}                     & \multirow{2}{*}{OC-SVM}          & Feature Cross           & 84.04\%           & 26.42\%         & 97.56\%          & 74.07\%            & 0.84              \\ \cline{3-8} 
                                         &                                  & Feature Concatenation   & 42.55\%           & 100.00\%         & 97.56\%          & 43.01\%            & 0.59              \\ \cline{2-8} 
                                         & \multirow{2}{*}{IF}              & Feature Cross           & 61.70\%           & 67.92\%         & 100.00\%          & 53.25\%            & 0.69              \\ \cline{3-8} 
                                         &                                  & Feature Concatenation   & 42.55\%           & 100.00\%         & 97.56\%         & 43.01\%            & 0.60              \\ \cline{2-8} 
                                         & \multirow{2}{*}{LOF}             & Feature Cross           & 69.15\%           & 54.72\%         & 100.00\%         & 58.57\%            & 0.74              \\ \cline{3-8} 
                                         &                                  & Feature Concatenation   & 43.62\%           & 100.00\%         & 100.00\%         & 43.62\%            & 0.61              \\ \hline
\end{tabular}
\label{performance}
\end{table*}

% \subsection{Performance of PUPGUARD under Different Conditions}
\subsection{Performance of PUPGUARD}
Table \ref{performance} presents the experimental results of the PUPGUARD method under different conditions. It is worth noting that the preprocessing method, Prepro2, mentioned earlier, is only combined with ResNet-based features, while Prepro1 is only combined with LBP- and HOG-based features.

Four types of deep learning-based features are evaluated using three classifiers, along with two feature fusion methods.
LBP- and HOG-based features are evaluated with the same classifiers.
It is noteworthy that regardless of which of the above feature extraction methods is used, we perform feature selection and dimensionality reduction on the extracted image features.

The methods using LBP- or HOG-based features for detecting puppet attacks demonstrate poor performance.% in behavior-based two-factor authentication. 
Regardless of the one-class classifier or feature fusion method employed, the best achieved performance is only 88.29\% accuracy and 15.09\% FPR. These results are insufficient for effective security defense.

In contrast, employing ResNet-based features significantly improves performance. Specifically, using ResNet50-based features, OC-SVM, and feature cross-fusion, PUPGUARD achieves the highest accuracy of 97.87\% and an FPR of 1.89\%.

Furthermore, under the premise of using ResNet features, feature cross-fusion outperforms feature concatenation noticeably. This can be attributed to our defined behavior patterns having one-dimensional timing characteristics, while image features exist in a high-dimensional space.

If solely employing feature concatenation to construct fused feature vectors, certain limitations and challenges arise. A significant limitation is the dimensionality mismatch between timing and image features, potentially leading to suboptimal performance by not fully utilizing their complementary information. Additionally, differences in feature scales could result in biased performance, favoring one feature type over others during the learning process.

In contrast, employing the feature cross-fusion method creates a more integrated and informative representation. Leveraging the inherent relationships between different feature types and their complementary strengths leads to improved performance and more accurate detection of puppet attacks. Moreover, feature cross-fusion mitigates dimensionality mismatch issues and ensures a more efficient and effective use of the combined feature set in the learning process.

\subsection{Detection Solely Based on Image Features}
The purpose of this experiment is to demonstrate the necessity of using both image features and timing characteristics in the PUPGUARD method to characterize our defined behavior patterns, in other words, to demonstrate the superiority of combining timing characteristics to detect puppet attacks.
Using only image features means that image features do not need to be fused with timing characteristics but are directly fed into a one-class classifier.

The performance of this experiment is shown in Table \ref{img_only_exp}.
The overall performance of this experiment is quite poor, with the highest achievable accuracy falling below 70\%, and the FPR is unacceptably high. This may be attributed to the following reasons: when coerced, the victim will make different degrees of resistance. When the victim's resistance is very strong, although the time interval between the two presses is much longer than normal, the force of pressing the fingerprint collection module may be normal or even too small due to resistance. In other words, in this case, the image features are normal but the timing characteristics is abnormal. If only the image features are used for puppet attack detection, there will be a high error rate and FPR.

\begin{table}[t]
\centering
\renewcommand\arraystretch{1.2}
\caption{Experimental Results Solely Based on Image Features}
\begin{tabular}{|l|l|c|c|}
\hline
\textbf{Features}                        & \textbf{Classifier} & \multicolumn{1}{l|}{\textbf{Accuracy}} & \multicolumn{1}{l|}{\textbf{FPR}} \\ \hline
\multirow{3}{*}{ResNet50} & OC-SVM     & 62.77\%                       & 66.04\%                  \\ \cline{2-4} 
                                & IF         & 64.89\%                       & 58.49\%                  \\ \cline{2-4} 
                                & LOF        & 46.81\%                       & 92.45\%                  \\ \hline
\multirow{3}{*}{LBP}            & OC-SVM     & 43.62\%                       & 98.11\%                  \\ \cline{2-4} 
                                & IF         & 43.62\%                       & 98.11\%                  \\ \cline{2-4} 
                                & LOF        & 43.62\%                       & 98.11\%                  \\ \hline
\multirow{3}{*}{HOG}            & OC-SVM     & 42.55\%                       & 100.00\%                 \\ \cline{2-4} 
                                & IF         & 43.62\%                       & 100.00\%                 \\ \cline{2-4} 
                                & LOF        & 43.62\%                       & 100.00\%                 \\ \hline
\end{tabular}
\label{img_only_exp}
\end{table}

\subsection{Detection Solely Based on timing characteristics}
The purpose of this experiment is to demonstrate the necessity of using both image features and timing characteristics in the PUPGUARD method to characterize our defined behavior patterns, in other words, to demonstrate the superiority of combining image features to detect puppet attacks. 
In this experiment, the input feature vector is only the timing characteristics, that is, the input is only one-dimensional features. The performance of this experiment is shown in Table \ref{timing_only}.

\begin{table}[t]
\centering
\renewcommand\arraystretch{1.2}
\caption{Experimental Results Solely Based on timing characteristics}
\begin{tabular}{|l|l|c|c|}
\hline
\textbf{Features}                & \textbf{Classifier} & \multicolumn{1}{l|}{\textbf{Accuracy}} & \multicolumn{1}{l|}{\textbf{FPR}} \\ \hline
\multirow{3}{*}{Timing} & OC-SVM     & 88.29\%                       & 11.32\%                   \\ \cline{2-4} 
                        & IF         & 89.36\%                       & 13.21\%                   \\ \cline{2-4} 
                        & LOF        & 89.36\%                       & 13.21\%                   \\ \hline
\end{tabular}
\label{timing_only}
\end{table}
% 本实验的性能比仅使用图像特征的实验要好，但是相较同时使用两种特征进行检测的方法，仍存在较大的性能差异。
% The overall performance of this experiment is also bad, the highest accuracy rate that can be achieved is slightly higher than 60\%, and the FPR is unacceptable. 
The performance of this experiment is better than the experiment using only image features, but there is still a large performance difference compared to the method that uses both features for detection.
This method also has obvious disadvantages, resulting in mediocre performance. Contrary to what was described in the previous subsection, in this case the attacker may have such a large power gap to the victim that the victim has to perform two quick presses. In this case, the time interval between pressings may be within the normal range, but the two pressing speeds are too fast and the force is too strong, resulting in excessive grayscale of the fingerprint image, severe deviation of the pressing center, or serious dragging marks in the pressing image. In other words, in this case, the image is abnormal but the timing characteristics is normal. If only the timing characteristics is used for detection, it will lead to huge risks. 

\begin{table}[t]
\centering
\renewcommand\arraystretch{1.2}
\caption{Experimental Results Using Decision Fusion}
\begin{tabular}{|c|c|c|c|}
\hline
\textbf{\begin{tabular}[c]{@{}l@{}}Image \\ Features\end{tabular}} & \textbf{\begin{tabular}[c]{@{}l@{}}Timing \\ Features\end{tabular}} & \textbf{Accuracy} & \textbf{FPR} \\ \hline
OC-SVM                                                             & OC-SVM                                                                     & 94.68\%           & 0.00\%       \\ \hline
OC-SVM                                                             & IF                                                                         & 95.74\%           & 1.89\%       \\ \hline
OC-SVM                                                             & LOF                                                                        & 95.74\%           & 1.89\%       \\ \hline
IF                                                                 & OC-SVM                                                                     & 92.55\%           & 1.89\%       \\ \hline
IF                                                                 & IF                                                                         & 92.55\%           & 3.77\%       \\ \hline
IF                                                                 & LOF                                                                        & 92.55\%           & 3.77\%       \\ \hline
LOF                                                                & OC-SVM                                                                     & 90.43\%           & 5.66\%       \\ \hline
LOF                                                                & IF                                                                         & 91.49\%           & 7.55\%       \\ \hline
LOF                                                                & LOF                                                                        & 91.49\%           & 7.55\%       \\ \hline
\end{tabular}
\label{decision_fusion}
\end{table}

\subsection{Performace using Decision Level Fusion}
% 我们也尝试不使用特征融合,而是使用两个分类器分别处理PUPGUARD中的两类特征，再使用决策融合处理两个分类器的分类结果，得到最终检测结果。
Instead of using feature fusion, we also try to use two classifiers to process the two types of features in PUPGUARD separately, and then use decision fusion to process the classification results of the two classifiers to get the final detection results.
% 因此检测傀儡攻击的任务被分解为两个子任务,即判断指纹图像是否合法,以及判断按压间隔是否合法.
Thus the task of detecting the puppet attack is decomposed into two subtasks, i.e., determining whether the fingerprint image is legitimate or not, and determining whether the press interval is legitimate or not.
% 根据上面两个实验的实验结果,我们使用基于ResNet50的特征作为图像特征。
Based on the experimental results of the above two experiments,we use ResNet50-based features as image features.
% 具体而言，我们使用一个单类分类器判别图像特征，同时使用一个单类分类器判别时间特征。之后,我们对两个分类器返回的分类结果进行逻辑与,即当且仅当两个分类器的判断结果都为合法用户时,最终的决策是用户合法.
Specifically, we use a one-class classifier to discriminate image features and a one-class classifier to discriminate timing characteristics at the same time.
After that, we use the logical ``and" relationship to process the classification results returned by the two classifiers, and the final decision is that the user is legitimate if and only if the results of both classifiers are normal.

% 使用决策融合的实验结果如图所示.
The experimental results using decision fusion are shown in Table \ref{decision_fusion}.
As can be seen from the experimental results, the overall performance using decision layer fusion is good. It can be noted that FPR that can be achieved with decision fusion is generally very low, even reaching 0.00\% at one point. This is due to the fact that we use the logical ``and" operation in decision fusion.
However, it can be seen that the accuracy of this method is not as good as the method of feature fusion used in PUPGUARD, which is due to the fact that the method of decision fusion produces too many FN values.

\subsection{Detection with Same Finger Pressed Twice}
The purpose of this experiment is to demonstrate the necessity of using two different fingers in PUPGUARD. Specifically, in constructing the dataset, volunteers were asked to use the same finger to press twice, with the same requirements as described in Section 4. To complete this experiment, we invited the same volunteers as those who created the dataset described in Section 4, and each person completed two presses using the thumb, finger, middle finger, ring finger and little finger respectively, collecting a total of 282 fingerprint pairs and time interval data as the training dataset. At the same time, we also created a test dataset using the method described in subsection 1 of this chapter, which includes 50 fingerprint pairs and time interval data.

\begin{table}[t]
\centering
\renewcommand\arraystretch{1.2}
\caption{Experimental Results with Same Finger Pressing Twice}
\begin{tabular}{|l|l|l|c|c|}
\hline
\textbf{Features}                  & \textbf{Classifier}              & \begin{tabular}[c]{@{}l@{}}\textbf{Feature}\\ \textbf{Fusion}\end{tabular} & \multicolumn{1}{l|}{\textbf{Accuracy}} & \multicolumn{1}{l|}{\textbf{FPR}} \\ \hline
\multirow{6}{*}{ResNet50} & \multirow{2}{*}{OC-SVM} & Cross                                                    & 76.00\%                       & 33.33\%                  \\ \cline{3-5} 
                          &                         & Concatenation                                            & 76.00\%                       & 33.33\%                  \\ \cline{2-5} 
                          & \multirow{2}{*}{IF}     & Cross                                                    & 88.00\%                       & 25.00\%                  \\ \cline{3-5} 
                          &                         & Concatenation                                            & 80.00\%                       & 41.67\%                  \\ \cline{2-5} 
                          & \multirow{2}{*}{LOF}    & Cross                                                    & 68.00\%                       & 66.67\%                  \\ \cline{3-5} 
                          &                         & Concatenation                                            & 60.00\%                       & 83.33\%                  \\ \hline
\multirow{6}{*}{LBP}      & \multirow{2}{*}{OC-SVM} & Cross                                                    & 60.00\%                       & 83.33\%                  \\ \cline{3-5} 
                          &                         & Concatenation                                            & 52.00\%                       & 100.00\%                 \\ \cline{2-5} 
                          & \multirow{2}{*}{IF}     & Cross                                                    & 60.00\%                       & 83.33\%                  \\ \cline{3-5} 
                          &                         & Concatenation                                            & 52.00\%                       & 100.00\%                 \\ \cline{2-5} 
                          & \multirow{2}{*}{LOF}    & Cross                                                    & 60.00\%                       & 83.33\%                  \\ \cline{3-5} 
                          &                         & Concatenation                                            & 52.00\%                       & 100.00\%                 \\ \hline
\multirow{6}{*}{HOG}      & \multirow{2}{*}{OC-SVM} & Cross                                                    & 56.00\%                       & 83.33\%                  \\ \cline{3-5} 
                          &                         & Concatenation                                            & 48.00\%                       & 100.00\%                 \\ \cline{2-5} 
                          & \multirow{2}{*}{IF}     & Cross                                                    & 60.00\%                       & 83.33\%                  \\ \cline{3-5} 
                          &                         & Concatenation                                            & 52.00\%                       & 100.00\%                 \\ \cline{2-5} 
                          & \multirow{2}{*}{LOF}    & Cross                                                    & 60.00\%                       & 83.33\%                  \\ \cline{3-5} 
                          &                         & Concatenation                                            & 52.00\%                       & 100.00\%                 \\ \hline
\end{tabular}
\label{SF}
\end{table}

\begin{figure}[h]
\centering
\subfloat[Two fingerprints pressed twice using the same finger.]{
    \begin{minipage}{0.49\linewidth}
		\vspace{2pt}		\centerline{\includegraphics[width=24mm,height=24mm]{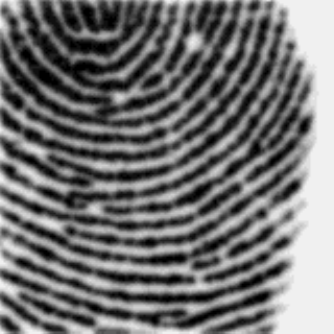}}
	\end{minipage}
	\begin{minipage}{0.49\linewidth}
		\vspace{2pt}		\centerline{\includegraphics[width=24mm,height=24mm]{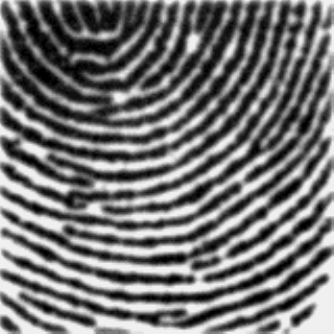}}
	\end{minipage}
 }
 
\subfloat[Two fingerprints pressed in succession by two different fingers.]{
    \begin{minipage}{0.49\linewidth}
		\vspace{2pt}		\centerline{\includegraphics[width=24mm,height=24mm]{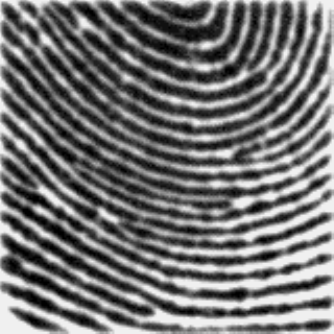}}
	\end{minipage}
	\begin{minipage}{0.49\linewidth}
		\vspace{2pt}		\centerline{\includegraphics[width=24mm,height=24mm]{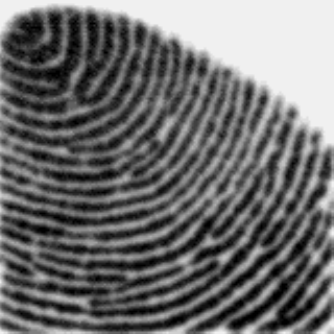}}
	\end{minipage}
 }
\caption{Comparison of fingerprint pairs using the same finger and different fingers.}
\label{fingerprint_com}
\end{figure}

It can be seen that this method has very obvious flaws, namely, a high FPR and low accuracy. The reason for this is related to the way the pressings are done. When the user needs to press two different fingers in succession, there must be a finger-switching action, which will cause significant changes in the angle, press center, and press intensity of the two presses. In this experiment, the user only needs to press the same finger twice in a row, and almost all users only lift their finger slightly after the first press to complete the second press, which will result in the fingerprint images of the two presses being extremely similar. As shown in Fig. \ref{fingerprint_com}, Fig. \ref{fingerprint_com}(a) shows two fingerprints pressed twice using the same finger, while Fig. \ref{fingerprint_com}(b) shows two fingerprints pressed in succession by two different fingers. It can be clearly seen from the Fig. \ref{fingerprint_com}(a) that the two fingerprints are almost the same. Therefore, in this case, the data in the data set cannot include all the pressed fingerprints under normal conditions. In other words, when the input positive samples are too limited, the hyperplane output by the model deviates greatly from the actual hyperplane, resulting in lower accuracy, lower precision and higher FPR. Moreover, from a practical point of view, this verification method will reduce the attack difficulty of the attacker, because the attacker does not need to force the victim to switch fingers, but only needs to forcibly lift the victim's finger and then press the fingerprint module. The performance of this experiment is shown in Table \ref{SF}.

\subsection{Effect of Dataset Size on PUPGUARD Performance}
The previous experiments have already demonstrated that using ResNet50 features and feature cross outperforms other methods. Therefore, when exploring the impact of the dataset size on PUPGUARD, we will only focus on using ResNet50 features and feature cross.

\begin{figure}[htb]
\centering
\includegraphics[width=84mm]{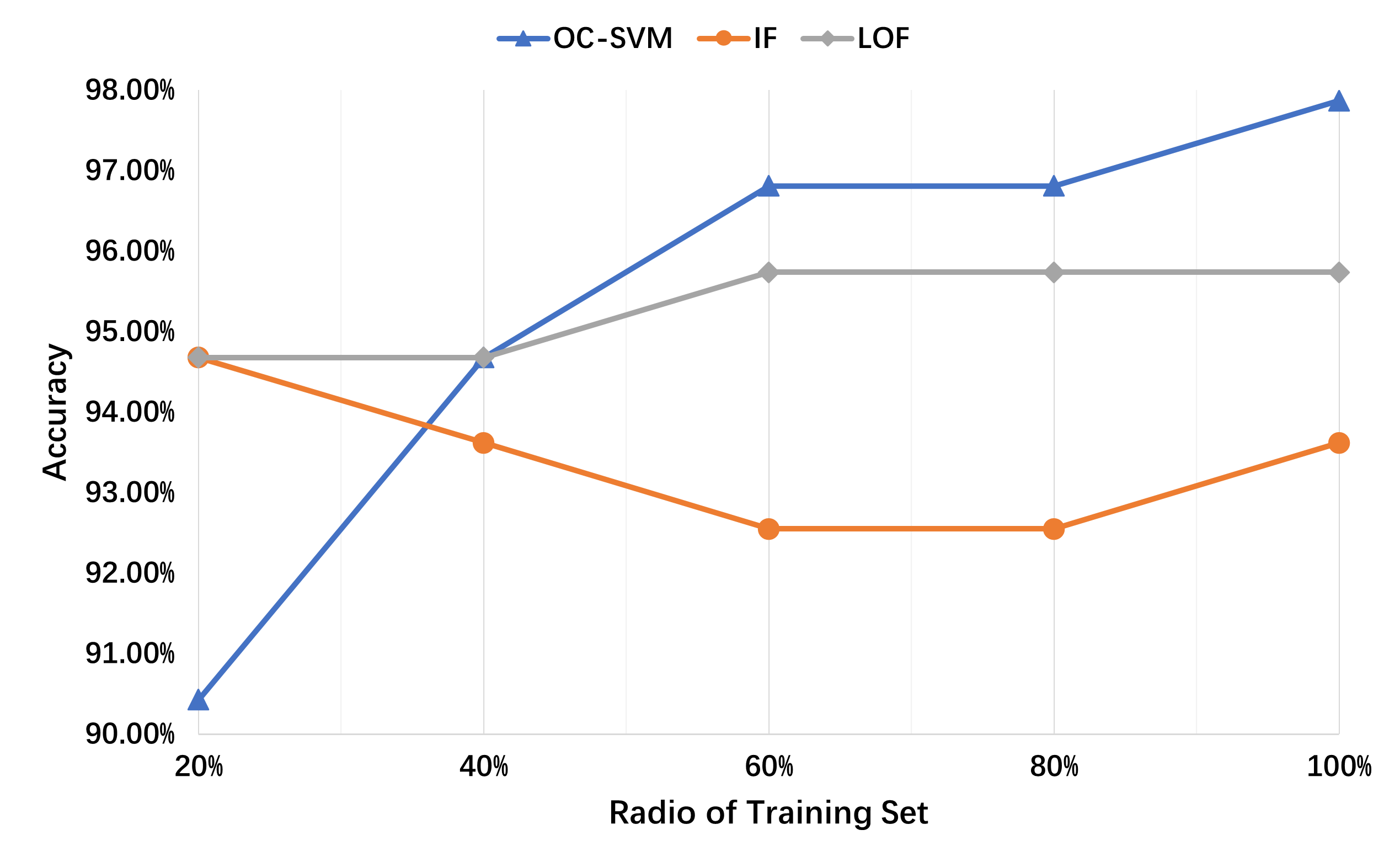}
\caption{Performance of PUPGUARD at different dataset sizes.}
\label{acc}
\end{figure}

To explore the effect of training dataset size on detection performance, we use 20\%, 40\%, 60\%, 80\%, and 100\% of the training dataset for training, respectively. 
Fig. \ref{acc} shows the impact of different dataset sizes on various detection performances. 

It can be observed that as the size of the traning set increases, the detection accuracy of PUPGUARD gradually stabilizes. In fact, the accuracy of OC-SVM and IF methods steadily improves. Therefore, we can draw the conclusion that the detection performance of PUPGUARD does improve as the training set increases.

\section{Limitations of PUPGUARD}
% \section{Limitations and Future Work}
% \subsection{Limitations of PUPGUARD}
\subsection{User Adoption and Usability} Requiring users to follow a specific sequence of actions, such as pressing the fingerprint module twice with distinct fingers, might result in resistance or confusion among users. The added steps could potentially lead to a decline in user adoption due to increased complexity, affecting the overall usability and user experience of the authentication process.

\subsection{Implementation and Technical Constraints} Implementing a behavior-based authentication approach like PUPGUARD might require adjustments to hardware, software, and user interfaces. Adapting existing authentication systems or developing new ones to incorporate dynamic behavior patterns can introduce technical challenges, compatibility issues, and potential vulnerabilities that must be carefully addressed to ensure the method's reliability and security.

\section{Conclusions}

In this paper, we present PUPGUARD, a solution crafted to provide protection against puppet attacks. PUPGUARD harnesses user behavior patterns, particularly the sequence of pressing the fingerprint module with different fingers, to capture inherent image features and timing characteristics. By adopting this two-factor authentication approach, we fortify security against puppet attacks, prioritizing the observation of dynamic behavior patterns throughout the authentication process. The requirement for two separate finger presses introduces an extra layer of security, with the time gap between these presses increasing the complexity for potential attackers. This comprehensive approach enhances security against fingerprint presentation attacks.

To evaluate the effectiveness of PUPGUARD, we performed experiments using datasets gathered from 31 subjects, encompassing both image features and timing characteristics. These data collection procedures were carried out with the approval of the Institutional Review Board (IRB). The results of our experiments clearly illustrate PUPGUARD's exceptional performance, achieving the highest accuracy at 97.87\% and the lowest false positive rate (FPR) at 1.89\%, respectively. Additionally, we conducted comparative experiments to affirm the advantage of incorporating both image features and timing characteristics into PUPGUARD, thereby reinforcing its resistance against puppet attacks.

\bibliographystyle{IEEEtran}
\bibliography{myref}

\end{document}